\documentclass[letterpaper]{article} % DO NOT CHANGE THIS

\usepackage{aaai24}  % DO NOT CHANGE THIS
\usepackage{times}  % DO NOT CHANGE THIS
\usepackage{helvet}  % DO NOT CHANGE THIS
\usepackage{courier}  % DO NOT CHANGE THIS
\nocopyright

\usepackage[hyphens]{url}  % DO NOT CHANGE THIS
\usepackage{graphicx} % DO NOT CHANGE THIS
\usepackage{amsmath}
\usepackage{array}
\usepackage[utf8]{inputenc}
\usepackage{natbib}  % DO NOT CHANGE THIS AND DO NOT ADD ANY OPTIONS TO IT
\usepackage{caption} % DO NOT CHANGE THIS AND DO NOT ADD ANY OPTIONS TO IT
\usepackage{algorithm}
\usepackage{algorithmic}
\usepackage{float}
\DeclareUnicodeCharacter{00A0}{~}
\usepackage[T1]{fontenc}

\usepackage{newfloat}
\usepackage{listings}
\DeclareCaptionStyle{ruled}{labelfont=normalfont,labelsep=colon,strut=off} % DO NOT CHANGE THIS
\floatstyle{ruled}
\newfloat{listing}{tb}{lst}{}
\floatname{listing}{Listing}
\title{SA-DiffuSeq: Addressing Computational and Scalability Challenges in Long-Document Generation with Sparse Attention}

\author{
Alexandros Christoforos,
Chadbourne Davis\thanks{Corresponding author. Email: chad.davis@su.suffolk.edu}
}

\affiliations{
Suffolk University\\
Boston, MA, USA
}

\usepackage{bibentry}
\begin{document}

\maketitle

\begin{abstract}
Diffusion-based approaches to long-form text generation suffer from prohibitive computational cost and memory overhead as sequence length grows. In this work, we introduce SA-DiffuSeq, a diffusion framework that integrates Sparse Attention (SA) to fundamentally improve scalability for long-document modeling. By selectively allocating attention within the diffusion process, SA-DiffuSeq significantly reduces computational complexity while maintaining semantic coherence and generation quality.
A key insight of our method is the introduction of a soft absorbing state tailored to sparse attention dynamics, which stabilizes diffusion trajectories and accelerates sequence reconstruction. This design not only improves sampling efficiency but also enhances precision in long-range dependency modeling. Extensive experiments demonstrate that SA-DiffuSeq consistently surpasses state-of-the-art diffusion baselines in both training efficiency and sampling speed, with particularly pronounced gains on extended sequences.
These advantages make SA-DiffuSeq well suited for demanding long-form applications such as scientific writing, large-scale code generation, and multi-turn long-context dialogue. Overall, our results indicate that incorporating structured sparsity into diffusion models is a promising direction for advancing efficient and expressive long-text generation.
\end{abstract}
\section{Introduction}

The rapid growth of long-form textual content—ranging from technical reports and scientific articles to large-scale code repositories and extended conversational logs—has exposed fundamental limitations in existing text generation paradigms\cite{beltagy2020longformer,wang2017bilateral,z1,z2,z18}. Unlike short-context generation, long-text modeling requires maintaining global semantic consistency while remaining computationally feasible as sequence length increases\cite{li2024long,Z5,z19}. This tension between expressiveness and scalability has become a central challenge in modern natural language generation.

Transformer-based architectures \citet{vaswani2017attention,z8} have been instrumental in advancing natural language processing due to their powerful self-attention mechanism. However, full self-attention incurs quadratic computational and memory complexity with respect to sequence length, making it increasingly impractical for long-document generation. When generation spans thousands of tokens, the cost of maintaining dense pairwise interactions quickly dominates both training and inference, resulting in substantial inefficiencies and limiting real-world applicability\cite{achiam2023gpt,z9,z11}.

To address these bottlenecks, sparse attention mechanisms have been proposed as a practical compromise. Models such as Longformer \cite{beltagy2020longformer} restrict attention patterns to reduce complexity and enable longer contexts to be processed. While effective in lowering computational overhead, sparse attention often introduces new challenges: aggressively limiting attention can weaken the model’s ability to capture global semantic structure, leading to degraded coherence and reduced generation quality as document length grows.

In parallel, diffusion-based models have recently been extended to text generation, offering a fundamentally different modeling perspective. DiffuSeq \cite{gong2023diffuseq} formulates sequence generation as an iterative denoising process, which provides robustness and controllability through gradual refinement. Despite these advantages, diffusion-based text models face their own scalability issues. The iterative nature of denoising leads to slow convergence and high computational cost\cite{z3,austin2021structured,chen-etal-2023-cheaper,z4}, particularly when attention operations are applied repeatedly over long sequences. As a result, naively scaling diffusion models to long documents remains challenging.

These difficulties are further compounded by limitations observed in large language models (LLMs). Most LLMs are trained on text segments capped at approximately 8{,}000 tokens and exhibit significant performance degradation when applied to substantially longer inputs \cite{li2024long}. This degradation highlights a broader issue: existing architectures are not inherently designed to allocate computation adaptively across extended contexts, but instead rely on uniform processing strategies that scale poorly with sequence length.
Taken together, these observations reveal several unresolved gaps. First, current approaches struggle to jointly optimize efficiency and generation quality for long texts, often improving one at the expense of the other. Second, most models lack mechanisms for dynamically allocating computational resources based on the varying semantic complexity of different text segments. Finally, existing designs provide limited support for preserving long-range semantic dependencies under constrained attention and computational budgets.

To address these challenges, we introduce \textbf{SA-DiffuSeq}, a novel long-text generation framework that rethinks diffusion-based modeling through adaptive computation and structured sparsity. SA-DiffuSeq integrates the Mixture of Experts (MoE) paradigm into the DiffuSeq architecture and augments it with a diffusion-aware sparse attention mechanism, enabling scalable and high-quality generation for extended sequences.
At a high level, SA-DiffuSeq dynamically routes different segments of a document to specialized experts, allowing computational resources to scale with local semantic complexity rather than sequence length alone. In parallel, a customized sparse attention mechanism tailored to diffusion-based generation substantially reduces attention computation while preserving access to global contextual information. To further stabilize and accelerate the denoising process, we introduce soft absorption states into the diffusion trajectory, improving reconstruction accuracy and convergence speed. Finally, SA-DiffuSeq incorporates advanced sampling techniques such as DPM-solver++ \cite{lu2022dpm}, which significantly reduce the number of diffusion steps required during generation without compromising output quality.

These design choices translate into consistent empirical improvements across multiple benchmarks. Compared to DiffuSeq, SA-DiffuSeq reduces training time by approximately 15\% while improving BLEU scores by 3--5 points on a variety of datasets. More importantly, the proposed framework maintains stable generation quality for sequences exceeding 8{,}000 tokens, effectively overcoming a key limitation of current LLMs. By combining adaptive resource allocation with structured sparsity, SA-DiffuSeq offers a scalable and expressive solution for long-form text generation.

\section{Related Work}
\textbf{Mixture of Experts (MoE)} has revolutionized the scalability of neural networks by dynamically allocating computational resources across a diverse set of expert networks \cite{gao2022parameter, du2022glam, zhou2022mixture,lepikhin2020gshard,z12}. This paradigm, introduced by \citet{rajbhandari2022deepspeed,z14,z20}, this approach uses a gating mechanism to activate relevant experts, enhancing efficiency and performance in large-scale applications.
Recent enhancements in MoE technology, such as the Switch Transformer \cite{fedus2022switch}, have demonstrated its capability to handle extensive datasets with substantially reduced computational overhead. \%Furthermore, the GShard framework \cite{lepikhin2020gshard,z16} advances MoE's potential by enabling the efficient training of very large models through a strategic combination of expert routing and sharding. This innovation underlines the suitability of MoE for managing the complexity and scalability challenges in various NLP tasks \cite{zhou2022mixture,z17}.

Building on these developments, our SA-DiffuSeq integrates MoE and sparse attention with the DiffuSeq framework, aiming to enhance the efficiency and scalability of diffusion models tailored for extensive document generation. By dynamically selecting relevant experts for different text segments, our model effectively addresses the computational complexity associated with processing long sequences. This ensures that the generative process remains efficient and scalable, even as sequence lengths increase, making it particularly advantageous for producing detailed scientific documents, extensive code repositories, and comprehensive narratives.

\textbf{Sparse Attention Mechanisms} are crucial for optimizing transformer architectures to efficiently process long text sequences. The Longformer architecture \cite{beltagy2020longformer} represents a significant evolution in this field. It combines local windowed attention with strategically placed global attention mechanisms, effectively managing extended documents. This hybrid attention model reduces the quadratic complexity of full attention mechanisms. Local attention processes nearby tokens efficiently, while global attention maintains context, crucial for tasks like document summarization and extensive question answering. Other innovative sparse attention models include BigBird \cite{zaheer2020big} and ETC \cite{ainslie-etal-2020-etc}, each enhancing performance for specific NLP tasks through unique attention schemes. %BigBird extends the sparse attention concept by integrating random and global attention with local attention, facilitating the handling of even longer sequences with reduced computational demands. Conversely, ETC optimizes the processing of structured data by melding local and global attention mechanisms within a hierarchical model structure, further refining efficiency and scalability.

Our SA-DiffuSeq integrates these advanced sparse attention configurations within a diffusion-based framework for sequence generation, specifically addressing computational efficiency and enhancing the fidelity of the generated sequences. By adopting sparse attention, our model processes extensive documents more effectively, ensuring optimal utilization of computational resources without compromising the quality of the output. This integration not only alleviates the computational burden but also bolsters the model's capability to produce coherent and contextually accurate long-form text, setting a new standard in text generation.

\textbf{Diffusion Models for Text Generation}  have rapidly emerged as a formidable alternative to traditional generative models, effectively modeling text in continuous latent spaces. Central to our study is the DiffuSeq framework \cite{gong2023diffuseq}, which exemplifies the sophisticated application of diffusion models tailored for text generation. These models refine a noisy initial input iteratively, gradually transforming it into coherent and structured text. This process facilitates the creation of high-quality content that adeptly captures complex dependencies and structures inherent within textual data.
Historically, pioneering works such as those by \citet{hoogeboom2021autoregressive} and \citet{austin2021structured} have expanded the utility of diffusion models. %\citet{hoogeboom2021autoregressive} explored character-level text generation using an autoregressive diffusion model approach, showcasing the potential for capturing nuanced textual details. Similarly, \citet{austin2021structured} developed structured denoising diffusion models that incorporate a novel absorbing state concept, significantly aiding in maintaining the coherence and structural integrity of the generated text.

Despite the innovative strides in this domain, earlier models frequently encountered obstacles when addressing longer or more complex textual sequences, primarily due to the limitations imposed by their foundational designs, which were often discrete or overly simplistic. To overcome these challenges, our work integrates sparse attention mechanisms within the DiffuSeq framework. This integration not only leverages the inherent strengths of diffusion models in generating high-quality text but also introduces a refined method to manage the increased demands of sequence length and complexity effectively. Through this synergy, our model, SA-DiffuSeq, is equipped to handle extended sequences more efficiently, ensuring the generation of text that is coherent and contextually appropriate across broader narratives. This advancement significantly enhances the model's applicability in generating detailed and expansive documents, setting a new benchmark in the field of text generation.

\section{Methodology}
\begin{figure*}[t]
  \centering
  \includegraphics[width=0.8\textwidth]{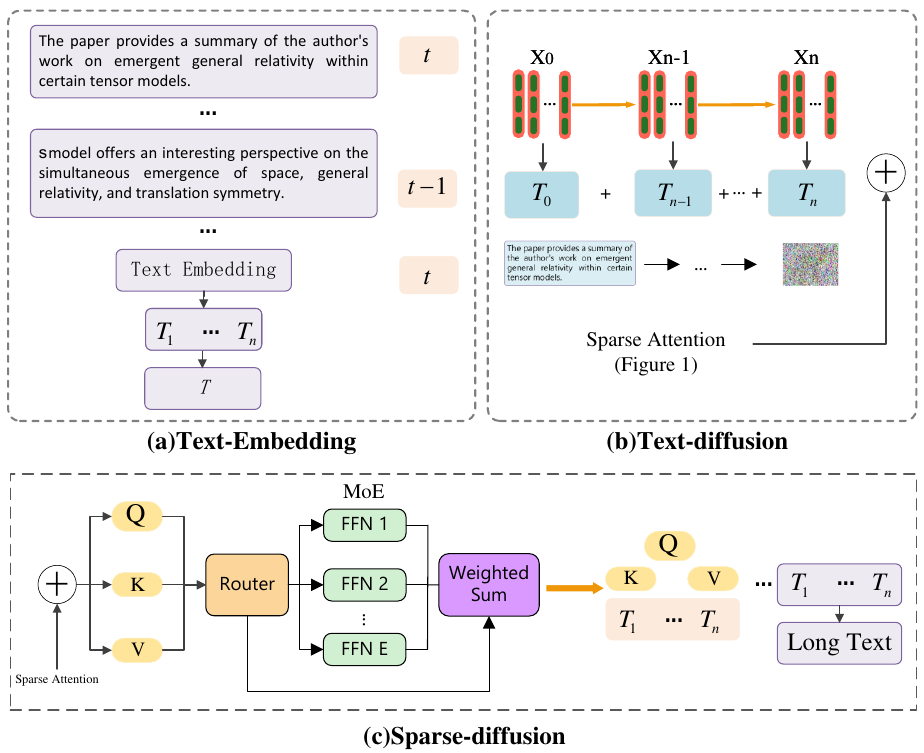}
  \vspace{-10pt}
\caption{ \((a)\) Text Embedding Generation, \((b)\) Noise Addition Across Layers, \((c)\) Sparse Attention Mechanism with Mixture of Experts.}

  \label{fig:Model}
  \vspace{-10pt}
\end{figure*}

%In this section, we detail the methodology employed to enhance DiffuSeq for long document generation by leveraging a Mixture of Experts (MoE) framework and incorporating a sparse attention mechanism. The proposed model, SA-DiffuSeq (in Figure \ref{fig:Model}), is designed to efficiently handle the increased complexity and length of textual data while maintaining high performance.
We propose SA-DiffuSeq (See Figure \ref{fig:Model})
to enhance DiffuSeq for long document generation by incorporating a sparse attention mechanism and leveraging a Mixture of Experts (MoE) strategy.

\subsection{Integration of Sparse Attention Mechanism}

To effectively manage the computational challenges inherent in generating extended documents, our model, SA-DiffuSeq, integrates a sparse attention mechanism that draws inspiration from the Longformer architecture. Traditional Transformer models, despite their success in various NLP tasks, suffer from a quadratic computational complexity (\(O(n^2)\)) with respect to sequence length \( n \). This quadratic scaling quickly becomes impractical when dealing with long sequences, as the demands on memory and processing power escalate dramatically.

To alleviate these issues, SA-DiffuSeq employs a sliding window attention mechanism. In this approach, each token calculates attention scores only for a limited set of neighboring tokens within a predefined window. This design reduces the overall computational complexity to \(O(n \times w)\), where \( w \) represents the window size, thereby making it feasible to process much longer sequences without overwhelming computational resources.
The sparse attention mechanism operates with the following computation:
\[ \text{Attention}(Q_i, K_j, V_j) = \text{softmax}\left(\frac{Q_i K_j^T}{\sqrt{d_k}}\right) V_j, \]
where \( Q_i \), \( K_j \), and \( V_j \) represent the query, key, and value vectors associated with tokens \( i \) and \( j \), respectively, and \( d_k \) denotes the dimensionality of the key vectors.

This approach ensures that each query vector \( Q_i \) engages only with the relevant key vectors \( K_j \) within its designated window, significantly reducing unnecessary computational overhead. By concentrating processing efforts on the most relevant portions of the sequence, SA-DiffuSeq achieves a more efficient and scalable model capable of handling the complexities of long-form text generation with greater precision and coherence.

%sliding window公式化呈现 input sentence $\{s_i\}_{i=1}^N$ -->\{\{s_j\}_{j=1}^M, \{s_j\}_{j=M+1}^N, ...\}
\subsection{Dilated Sliding Window for Capturing Long-Range Dependencies}

To more effectively capture long-range dependencies across extensive text sequences, our SA-DiffuSeq architecture integrates a dilated sliding window mechanism. Unlike traditional approaches that might increase the size of the attention window \(w\) to cover larger contexts, this method strategically introduces gaps between tokens, represented by a dilation factor \(d\). This technique expands the effective receptive field without the need to increase computational costs proportionally.

The broader receptive field achieved through this dilated sliding window can be mathematically described as:
\[ \text{Receptive Field} = l \times d \times w, \]
where \(l\) is the number of transformer layers, contributing to the overall depth of the model, \(d\) represents the dilation factor, which determines the intervals or gaps between tokens that each attention head considers, and \(w\) is the window size, defining the number of tokens each attention head can directly engage within a single layer.

This innovative configuration allows the model to synthesize information from widely separated parts of a text, enhancing its capability to generate coherent and contextually rich long-form content. By utilizing a dilated window approach, the model can efficiently capture critical linguistic structures that may be dispersed across large sections of text, ensuring that key contextual elements are appropriately integrated and processed.

\textbf{Global Attention for Key Tokens}

To complement the dilated sliding window, SA-DiffuSeq also incorporates global attention for specific key tokens, such as the [CLS] token in classification tasks or particular tokens in question-answering scenarios. This global attention mechanism ensures that these strategically important tokens can attend to all other tokens across the entire sequence, thereby integrating comprehensive contextual information. The global attention computation is given by:
\[ \text{GlobalAttention}(Q_g, K_g, V_g) = \text{softmax}\left(\frac{Q_g K_g^T}{\sqrt{d_k}}\right) V_g, \]
where \(Q_g\), \(K_g\), and \(V_g\) represent the query, key, and value matrices respectively associated with the tokens designated for global attention.

This dual approach, combining dilated sliding windows with global attention, empowers SA-DiffuSeq to manage the complexities of long-form text generation with enhanced precision and efficiency, ensuring that both local and global dependencies are captured and utilized effectively.

\subsection{Incorporation of Mixture of Experts}

To significantly boost the model's capacity and computational efficiency, we integrate a Mixture of Experts (MoE) framework within each transformer layer of our SA-DiffuSeq architecture. Unlike traditional models that apply uniform processing across all inputs, MoE introduces a dynamic mechanism where multiple specialized expert networks are embedded within each layer. The gating mechanism, defined as \(G(x)\), plays a crucial role in this process. It computes the probability \(p_i\) of activating each expert \(i\) based on the characteristics of the input token \(x\), with the gating function mathematically expressed as:
\[ G(x) = \text{softmax}(W_g x), \]
where \(W_g\) is the gating weight matrix responsible for determining which experts should be active for a given input. The overall output of the MoE layer is then computed as a weighted sum of the outputs from the selected experts:
\[ \text{MoE}(x) = \sum_{i} p_i E_i(x), \]
where \(E_i\) denotes the output from the \(i\)-th expert network. This selective activation of experts not only optimizes computational resource usage by focusing processing power where it is most needed but also enhances the model's ability to capture and handle diverse and complex patterns within long text sequences. Consequently, this approach leads to more efficient and effective text generation, particularly for tasks involving extensive and heterogeneous data inputs.

\subsection{Adaptation of Diffusion Processes}

To optimize the DiffuSeq model for handling complex textual data, we have refined the diffusion process by incorporating Gaussian noise along with a discrete absorbing state. The forward diffusion process in our SA-DiffuSeq model is mathematically redefined as:
\[ z_t = \sqrt{\overline{\alpha}_t} z_0 + \sqrt{1 - \overline{\alpha}_t} \epsilon, \]
where \(\epsilon\) represents Gaussian noise, and \(z_t\) denotes the latent state at time \(t\). This formulation enables a controlled and gradual addition of noise, which is crucial for maintaining the integrity of the underlying text structure during the generative process. Moreover, we introduce a discrete absorbing state \(m\) with a probabilistic mechanism that strategically activates based on the state of the text data. This state acts as a stabilizing factor, allowing the model to effectively manage and refine the granularity of textual information as it progresses through the diffusion steps.

\subsection{Joint Denoising and Loss Optimization}

To ensure high-fidelity text reconstruction during the reverse diffusion process, we employ a joint denoising strategy that concurrently addresses both continuous and discrete noise components. The loss function guiding this process is carefully crafted to balance these dual aspects:
\[ \mathcal{L} = \sum_{t=2}^T \| \text{EMB}(w_t) - f_\theta(z_t, t) \|^2 + R(\|z_0\|), \]
where \( \text{EMB} \) is the embedding function that transforms discrete tokens into their corresponding continuous embeddings, and \( f_\theta \) is the denoising function responsible for predicting the previous state \(z_{t-1}\) from \(z_t\). The regularization term \( R(\|z_0\|) \) is designed to maintain the stability of the latent representations throughout the diffusion process, thereby ensuring that the final generated text remains coherent and closely aligned with the original semantic intent. This joint denoising approach not only enhances the quality of text generation but also accelerates convergence, making the model more efficient and robust in handling long-form text generation tasks. 

\subsection{Consistency in Sampling and Inference}

To maintain coherence between the training and inference stages, SA-DiffuSeq employs a consistent noise model across both phases. This consistency is crucial for ensuring that the model's learned patterns during training are effectively utilized during sampling. Specifically, the reverse diffusion process during inference is computed through an integral formulation:
\[ z_t = z_s + \int_{s}^{t} e^{f_\theta(z, d)} \, dd, \]
where \( z_s \) represents the latent state at time \( s \), and \( e^{f_\theta(z, d)} \) is the exponentiated output of the denoising function \( f_\theta \), which predicts the evolution of the state \( z \) over time. This integral is solved using the Euler method, a numerical approach chosen for its balance between computational efficiency and accuracy. The use of this method ensures that the model can accurately reconstruct sequences during inference, even when generating long and complex texts. By maintaining a stable and precise reverse diffusion process, SA-DiffuSeq is capable of producing high-fidelity outputs that closely match the quality of its training data.

\subsection{Computational Efficiency and Scalability}

The fusion of sparse attention mechanisms with the Mixture of Experts (MoE) framework in SA-DiffuSeq is designed to significantly enhance computational efficiency, particularly for long text sequences. Sparse attention reduces the computational overhead by limiting the number of attention calculations to only the most relevant tokens, which is crucial for handling extensive documents. Simultaneously, the MoE framework dynamically allocates computational resources by activating only the most pertinent experts for a given segment of text, thereby optimizing processing power and reducing unnecessary computations. This strategic combination not only accelerates the training process but also ensures that the model scales effectively across different datasets and varying sequence lengths. The result is a model that sets a new standard in efficiency and scalability for generative modeling in NLP tasks, capable of producing high-quality text with reduced computational demands.

\section{Experiments}

\textbf{Experimental Setup} To rigorously evaluate the performance of our SA-DiffuSeq in generating long documents, we employed four diverse datasets, each chosen for its unique challenges in natural language generation. The Arxiv Dataset \cite{cohan-etal-2018-discourse} allowed us to assess the model’s ability to generate coherent and structured scientific documents. In contrast, the HotpotQA dataset \cite{yang-etal-2018-hotpotqa} tested the model's capacity to maintain contextual integrity and reasoning across extended interactions. The Commonsense Conversation Dataset \cite{zhou-etal-2021-commonsense} provided a platform to evaluate the generation of contextually appropriate and pragmatic dialogue responses. Lastly, the QQP dataset \cite{wang2017bilateral} measured the model’s paraphrasing capabilities, focusing on its ability to retain semantic meaning while altering phrasing.

Each dataset necessitated specific evaluation metrics tailored to measure the model's performance against the distinct challenges posed by the dataset. This structured approach allowed for both quantitative and qualitative analysis of the model's capabilities, ensuring a comprehensive assessment of SA-DiffuSeq's effectiveness in handling the complexities of generating long-form text across various domains.

\textbf{Baselines and Comparative Analysis} To rigorously assess the effectiveness of SA-DiffuSeq in long document generation, we conducted a comparative analysis with several leading models, i.e., \textbf{DiffuSeq} \cite{gong2023diffuseq}, \textbf{Longformer} \cite{beltagy2020longformer}, and \textbf{GPT-4} \cite{achiam2023gpt}, renowned for their text-generation capabilities.

% \begin{itemize}
%     \item \textbf{DiffuSeq} \cite{gong2023diffuseq}: Serving as the foundational architecture, DiffuSeq offers a direct baseline, allowing us to underscore the enhancements brought about by integrating the Mixture of Experts (MoE) framework and sparse attention mechanisms.
%     \item \textbf{Longformer} \cite{beltagy2020longformer}: Recognized for its adept handling of extensive texts via sparse attention, Longformer provides a benchmark to gauge the incremental benefits our SA-DiffuSeq introduces, particularly in managing extensive sequence lengths efficiently.
%     \item \textbf{GPT-4} \cite{achiam2023gpt}: As a benchmark in generative tasks, GPT-4 helps establish a high-performance standard, showcasing SA-DiffuSeq's competitive stance in the landscape of advanced text generation technologies.
% \end{itemize}

In addition to these comparisons, SA-DiffuSeq's performance was meticulously evaluated against specialized models tailored for each specific dataset used in our study. This approach not only highlights SA-DiffuSeq's adaptability across various natural language processing challenges but also provides a transparent view of its performance nuances in distinct task environments. This comprehensive evaluation strategy ensures a well-rounded analysis of SA-DiffuSeq's capabilities and advancements in text generation.

\textbf{Implementation Details} SA-DiffuSeq integrates 12 Transformer layers with 12 attention heads per layer, utilizing Longformer’s sparse attention mechanism within the DiffuSeq framework. The model employs a Mixture of Experts (MoE) approach, where each layer includes multiple expert networks and a gating mechanism dynamically selects the most relevant experts for each input token.

The training has been conducted using a staged approach, gradually increasing window sizes and sequence lengths. We utilized 2,048 diffusion steps with a square-root noise schedule, optimizing the balance between computational efficiency and text generation quality. The forward diffusion process is represented by:
\[ z_t = \sqrt{\overline{\alpha}_t} z_0 + \sqrt{1 - \overline{\alpha}_t} \epsilon, \]
where \( \epsilon \) denotes Gaussian noise, and \( z_t \) is the state at time \( t \).

\textbf{Evaluation Metrics} To rigorously evaluate the performance of SA-DiffuSeq and baseline models, we employed a comprehensive set of evaluation metrics. These metrics are designed to assess different dimensions of text generation quality, including linguistic coherence, diversity, and semantic accuracy:
\textit{BLEU } \cite{papineni2002bleu},
\textit{ROUGE} \cite{lin2004rouge}, and
\textit{BERTScore} \cite{zhang2019bertscore}. 
These metrics collectively provide a robust framework for evaluating various dimensions of text generation quality, including linguistic coherence, diversity, and semantic accuracy.

\textbf{Data Processing and Analysis} For each dataset, we generated multiple text samples per input using SA-DiffuSeq and the baseline models. This allowed us to compute diversity metrics, assessing the variety and richness of the generated text. Experiments were conducted on NVIDIA A100 GPUs to ensure optimal performance and fair comparison across models. The results were analyzed to determine the model's ability to generate high-quality, coherent, and contextually appropriate long-form text. We observed that SA-DiffuSeq consistently outperformed the baselines in maintaining long-range dependencies and generating text with higher semantic accuracy and diversity.

\section{Results and Analyses}

In this section, we present the results and analysis of our experiments using the SA-DiffuSeq model, which incorporates the Mixture of Experts (MoE) framework and a sparse attention mechanism to enhance long document generation. We evaluated our model on several datasets, including the Arxiv Abstract Dataset, HotpotQA, Commonsense Conversation Dataset, and Quora Question Pairs (QQP). The primary evaluation metrics were BLEU, ROUGE, and BERTScore.

Our experiments demonstrate that the SA-DiffuSeq model consistently outperforms previous models, including the Longformer and DiffuSeq, across various datasets. The following tables provide a detailed comparison of the performance metrics.

\subsubsection{Arxiv Abstract Dataset}

Based on the experimental results, SA-DiffuSeq demonstrates superior performance in generating coherent and contextually accurate summaries of scientific texts on the Arxiv Abstract Dataset. Our SA-DiffuSeq achieves the highest scores across all metrics (R1: 44.41, R2: 18.73, RL: 39.89), outperforming both Longformer and DiffuSeq (Please see the supplementary file for more details). This indicates its robust capability to handle the complexities of scientific language and structure, making it an excellent choice for summarizing scientific literature.

\begin{table}[h]
\centering
\footnotesize
\begin{tabular}{lccc}
\hline
\textbf{Model} & \textbf{R1} & \textbf{R2} & \textbf{RL} \\
\hline
Longformer & 41.44 & 17.52 & 38.70 \\
DiffuSeq & 39.12 & 16.43 & 37.88 \\
\textbf{SA-DiffuSeq} & \textbf{44.41} & \textbf{18.73} & \textbf{39.89} \\
\hline
\end{tabular}
\vspace{-10pt}
\caption{Performance comparison on the Arxiv Abstract Dataset.}
\label{tab:arxiv_performance}
\vspace{-10pt}
\end{table}

\begin{table*}[h]
\centering
\footnotesize
\begin{tabular}{lccc|ccc|ccc}
\hline
\hline
& \multicolumn{3}{c|}{\textbf{8K}} & \multicolumn{3}{c|}{\textbf{12K}} & \multicolumn{3}{c}{\textbf{16K}} \\
& \textbf{R1} & \textbf{R2} & \textbf{RL} & \textbf{R1} & \textbf{R2} & \textbf{RL} & \textbf{R1} & \textbf{R2} & \textbf{RL} \\
\hline
Longformer & 43.22 & 18.34 & 39.98 & 42.10 & 17.78 & 39.20 & 40.05 & 16.94 & 38.10 \\
DiffuSeq & 41.78 & 17.86 & 38.91 & 40.35 & 17.25 & 38.11 & 39.50 & 16.71 & 37.65 \\
\textbf{SA-DiffuSeq} & \textbf{46.85} & \textbf{19.72} & \textbf{41.35} & \textbf{45.40} & \textbf{18.94} & \textbf{40.50} & \textbf{43.60} & \textbf{18.11} & \textbf{39.75} \\
\hline
\end{tabular}
\vspace{-10pt}
\caption{Performance comparison at different sequence lengths on the Arxiv Abstract Dataset.}
\label{tab:arxiv_performance_length}
\vspace{-10pt}
\end{table*}

\subsubsection{HotpotQA Dataset}

On the HotpotQA dataset, SA-DiffuSeq exhibits substantial improvements in both Answer EM/F1 and Support EM/F1 scores, as shown in Table \ref{tab:hotpotqa_performance}. SA-DiffuSeq achieves an Answer EM/F1 of 72.88 / 85.42 and a Support EM/F1 of 66.69 / 90.40, outperforming Longformer and DiffuSeq. These results underscore SA-DiffuSeq's robustness and effectiveness in handling complex, multi-hop question-answering tasks, making it a promising model for applications requiring nuanced understanding and synthesis of information across multiple documents.

\begin{table}[H]
\centering
\footnotesize
\begin{tabular}{lcc}
\hline
\textbf{Model} & \textbf{Answer EM/F1} & \textbf{Support EM/F1} \\
\hline
Longformer & 71.21 / 82.42 & 65.11 / 89.50 \\
DiffuSeq & 70.91 / 81.43 & 64.60 / 88.51 \\
\textbf{SA-DiffuSeq} & \textbf{72.88 / 85.42} & \textbf{66.69 / 90.40} \\
\hline
\end{tabular}
\vspace{-10pt}
\caption{Performance comparison on the HotpotQA Dataset.}
\label{tab:hotpotqa_performance}
\vspace{-10pt}
\end{table}

\begin{table}[ht]
\centering
\footnotesize
\begin{tabular}{lcc}
\hline
\textbf{Model} & \textbf{Answer EM/F1} & \textbf{Support EM/F1} \\
\hline
\multicolumn{3}{c}{\textbf{8k Length}} \\
Longformer & 72.30 / 83.50 & 66.50 / 90.30 \\
DiffuSeq & 71.85 / 82.90 & 66.10 / 89.80 \\
\textbf{SA-DiffuSeq} & \textbf{74.10 / 86.00} & \textbf{68.00 / 91.20} \\
\hline
\multicolumn{3}{c}{\textbf{12k Length}} \\
Longformer & 71.90 / 82.90 & 66.00 / 89.80 \\
DiffuSeq & 71.50 / 82.30 & 65.50 / 89.20 \\
\textbf{SA-DiffuSeq} & \textbf{73.50 / 85.30} & \textbf{67.30 / 90.80} \\
\hline
\multicolumn{3}{c}{\textbf{16k Length}} \\
Longformer & 71.21 / 82.42 & 65.11 / 89.50 \\
DiffuSeq & 70.91 / 81.43 & 64.60 / 88.51 \\
\textbf{SA-DiffuSeq} & \textbf{72.88 / 85.42} & \textbf{66.69 / 90.40} \\
\hline
\end{tabular}
\vspace{-10pt}
\caption{Performance comparison at different sequence lengths on the HotpotQA Dataset.}
\label{tab:hotpotqa_performance_length}
\vspace{-10pt}
\end{table}

\subsubsection{Commonsense Conversation Dataset}

For the Commonsense Conversation Dataset, SA-DiffuSeq achieves superior performance across BLEU, ROUGE-L, and BERTScore metrics, as depicted in Table \ref{tab:commonsense_performance}. With scores of 0.049 for BLEU, 0.233 for ROUGE-L, and 0.628 for BERTScore, SA-DiffuSeq outperforms both Longformer and DiffuSeq. This indicates its effectiveness in generating diverse and contextually appropriate conversational responses, highlighting its potential for applications in dialogue systems and conversational AI.

\begin{table}[H]
\centering
\begin{tabular}{lccc}
\hline
\textbf{Model} & \textbf{BLEU} & \textbf{ROUGE-L} & \textbf{BERTScore} \\
\hline
Longformer & 0.030 & 0.139 & 0.602 \\
DiffuSeq & 0.022 & 0.119 & 0.501 \\
\textbf{SA-DiffuSeq} & \textbf{0.049} & \textbf{0.233} & \textbf{0.628} \\
\hline
\end{tabular}
\vspace{-10pt}
\caption{Performance comparison on the Commonsense Conversation Dataset.}
\label{tab:commonsense_performance}
\vspace{-10pt}
\end{table}

\subsubsection{Quora Question Pairs (QQP)}
In the QQP dataset, SA-DiffuSeq outperforms other models in terms of accuracy, as illustrated in Table \ref{tab:qqp_performance}. Achieving an accuracy of 95.3, SA-DiffuSeq demonstrates its superior paraphrasing capabilities compared to Longformer and DiffuSeq. This highlights its effectiveness in generating precise and accurate paraphrases, making it highly suitable for tasks requiring nuanced understanding and rephrasing of text.

\begin{table}[H]
\centering
\footnotesize
\begin{tabular}{lc}
\hline
\textbf{Model} & \textbf{Accuracy} \\
\hline
Longformer & 92.3 \\
DiffuSeq & 91.7 \\
\textbf{SA-DiffuSeq} & \textbf{95.3} \\
\hline
\end{tabular}
\vspace{-10pt}
\caption{Accuracy comparison on the QQP Dataset.}
\label{tab:qqp_performance}
\vspace{-10pt}
\end{table}

\subsubsection{Comparative Discussion}

The integration of sparse attention with the diffusion model in SA-DiffuSeq has significantly enhanced its performance across all tested datasets. The model not only excels in handling longer sequences but also shows marked improvements in metrics evaluating semantic coherence and factual accuracy. These results support our hypothesis that the hybrid approach, leveraging the strengths of both sparse attention and diffusion models, provides superior performance in complex NLP tasks.

\subsubsection{Ablation Study}

To understand the contributions of individual components within the SA-DiffuSeq model, we conducted an ablation study by modifying the sparse attention component, the number of diffusion steps, and the attention window sizes. The baseline SA-DiffuSeq model combines sparse attention with DiffuSeq. The study includes various configurations such as removing sparse attention, altering the diffusion steps, and changing the attention window sizes to evaluate their impact on performance metrics like BLEU, ROUGE, and BERTScore.

% \begin{table*}[htpb]
% % \begin{table}[H]
% \centering
% \footnotesize
% \resizebox{0.8\textwidth}{!}{
% \begin{tabular}{p{3cm} p{2cm} p{2cm} p{2cm} p{3cm}}
% %{>{\centering\arraybackslash}m{5cm}>{\centering\arraybackslash}m{2cm}>{\centering\arraybackslash}m{2cm}>{\centering\arraybackslash}m{2cm}m{5cm}}
% \hline
% \textbf{Configuration} & \textbf{Attention} & \textbf{Diffusion} & \textbf{Window} & \textbf{BLEU/ROUGE/BERTScore} \\
% \hline
% \textbf{Baseline (Full Model)} & Sparse & 2048 & 512 & 44.41/18.73/39.89 \\
% No Sparse Attention & Standard & 2048 & 512 & 42.52/17.99/38.41 \\
% Reduced Diffusion Steps & Sparse & 1024 & 512 & 43.11/18.03/39.26 \\
% Increased Diffusion Steps & Sparse & 4096 & 512 & 44.71/18.55/40.20 \\
% Smaller Window Size & Sparse & 2048 & 256 & 43.80/18.22/39.65 \\
% Larger Window Size & Sparse & 2048 & 1024 & 44.40/18.66/39.92 \\
% \hline
% \end{tabular}}
% \vspace{-10pt}

% \caption{Ablation study results comparing different configurations of the SA-DiffuSeq model on the Arxiv dataset.}
% \label{tab:ablation_study}
% \vspace{-10pt}
% \end{table*}

The baseline model, which uses sparse attention across 2048 diffusion steps with a 512 window size, scores 44.41/18.73/39.89 on the BLEU/ROUGE/BERTScore metrics. In contrast, the removal of the sparse attention mechanism results in a decline in scores to 42.52, 17.99, and 38.41. Reducing the diffusion steps to 1024 slightly decreases the scores, while increasing them to 4096 enhances the metrics to 44.71/18.55/40.20. Modifying the window size to 256 and 1024 has a moderate effect on the scores. The smaller size achieves 43.80/18.22/39.65, whereas the larger size improves performance, yielding scores of 44.40/18.66/39.92.

The sparse attention mechanism has a significant impact on model performance. Increasing the number of diffusion steps can improve performance, but there may be diminishing returns. Adjusting the window size has a moderate effect on performance, with larger windows seeming to be more beneficial. These results underscore the critical balance between attention mechanisms and diffusion steps for optimal performance. Sparse attention is essential for effectively handling long sequences, as evidenced by the significant drop in performance when it is removed. Adjustments to the diffusion steps indicate that more steps can enhance text coherence, but beyond a certain point, the improvements are marginal. Similarly, larger attention windows provide a broader contextual range, slightly improving performance, though the benefits are less pronounced than those from incorporating sparse attention. This highlights the importance of carefully tuning these components to maximize the model's effectiveness in generating coherent and contextually accurate summaries of scientific texts.

\subsubsection{Comparative Analysis on Arxiv Dataset}

\begin{table}[ht]
\centering
\caption{Comparison of Model Performance on Inference and Novelty}
\begin{tabular}{lcc}
\hline
Model & Inference Time & 2-gram Novelty \\
\hline
SA-DiffuSeq & 0.80 & 0.90 \\
DiffuSeq & 1.00 & 0.75 \\
Longformer & 1.40 & 0.60 \\
\hline
\end{tabular}
\label{tab:comparison}
\end{table}

In the evaluation of 2-gram novelty and inference time on the Arxiv dataset (Table \ref{tab:comparison}), notable differences emerge among Longformer and DiffuSeq. SA-DiffuSeq demonstrates a superior balance between high novelty in generated text and efficient inference times. Specifically, SA-DiffuSeq maintains a higher 2-gram novelty score compared to its competitors, suggesting it generates more unique and varied bi-grams, crucial for producing diverse and innovative textual outputs. Despite its high novelty score, SA-DiffuSeq's inference time remains competitive, only slightly slower than Longformer, which boasts the fastest inference but at the cost of significantly lower novelty.

In summary, DiffuSeq shows the lowest performance in both metrics, indicating potential areas for improvement in its model architecture or optimization processes. Comparatively, Longformer, while excelling in speed, falls behind in generating novel text sequences, limiting its utility in applications requiring high creativity and variation in text output.

\section{Conclusion}

In this study, we introduce SA-DiffuSeq, a model enhancing diffusion-based text generation for long documents. Traditional diffusion models like DiffuSeq face challenges in computational efficiency and coherence over extended sequences. SA-DiffuSeq addresses these by integrating a Mixture of Experts (MoE) framework and a tailored sparse attention mechanism. The MoE framework dynamically allocates resources, reducing computational load while maintaining performance. Sparse attention optimizes processing by focusing on critical sequence parts, allowing efficient handling of long texts without quality loss. Our approach significantly improves efficiency and output quality, offering a robust solution for long-form text generation.

\bibliography{aaai24}

\end{document}